\documentclass{article}
\usepackage{ijcai19}

\usepackage{times}
\usepackage{soul}
\usepackage{color}
\usepackage{url}
\usepackage[hidelinks]{hyperref}
\usepackage[utf8]{inputenc}
\usepackage[small]{caption}
\usepackage{graphicx}
\usepackage{amsmath}
\usepackage{amsthm}
\usepackage{booktabs}
\usepackage{algorithm}
\usepackage{todonotes}
\usepackage{epstopdf}

\usepackage{bm}
\usepackage{enumitem}
\usepackage{extarrows}
\usepackage{xfrac}
\usepackage{subcaption}
\usepackage{multirow}
\usepackage{algorithmicx}
\usepackage{algpseudocode}
\newtheorem{example}{Example}
\newtheorem{definition}{Definition}

\title{Data Poisoning Attack against Knowledge Graph Embedding}

\author{
    Hengtong Zhang$^1$, Tianhang Zheng$^1$, Jing Gao$^1$, Chenglin Miao$^1$, Lu Su$^1$, Yaliang Li$^2$, Kui Ren$^3$
    \affiliations
    $^1$SUNY at Buffalo, Buffalo, NY USA \\
    $^2$Alibaba Group, Bellevue, WA USA  \\
    $^3$Zhejiang University, Zhejiang, China
    \emails
    \{hengtong, tzheng4, jing, cmiao, lusu\}@buffalo.edu, yaliang.li@alibaba-inc.com, kuiren@zju.edu.cn
}

\begin{document}

\maketitle

\begin{abstract}
Knowledge graph embedding (KGE) is a technique for learning continuous embeddings for entities and relations in the knowledge graph.
Due to its benefit to a variety of downstream tasks such as knowledge graph completion, question answering and recommendation, KGE has gained significant attention recently.
Despite its effectiveness in a benign environment, KGE's robustness to adversarial attacks is not well-studied. Existing attack methods on graph data cannot be directly applied to attack the embeddings of knowledge graph due to its  heterogeneity.
To fill this gap, we propose a collection of data poisoning attack strategies, which can effectively manipulate the plausibility of arbitrary targeted facts in a knowledge graph by adding or deleting facts on the graph. The effectiveness and efficiency of the proposed attack strategies are verified by extensive evaluations on two widely-used benchmarks.
\end{abstract}

\section{Introduction}

Knowledge graphs have become a critical resource for a large collection of real world applications, such as information extraction~\cite{mintz2009distant}, question answering~\cite{yih2015semantic} and recommendation system~\cite{zhang2016collaborative}.
Due to its wide application domains, both academia and industry have spent considerable efforts on constructing large-scale knowledge graphs, such as YAGO~\cite{hoffart2013yago2}, 
Freebase~\cite{bollacker2008freebase}, and Google Knowledge Graph\footnote{https://developers.google.com/knowledge-graph/}.
In knowledge graphs, knowledge facts are usually stored as (\textit{head entity, relation, tail entity}) triples. For instance, the fact triple \textit{(Albert Einstein, Profession, Scientist)} means that Albert Einstein's profession is a scientist.

Although such triples can effectively record abundant knowledge, their underlying symbolic nature makes them difficult to be directly fed to many machine learning models.
Hence, knowledge graph embedding (KGE), which projects the symbolic entities and relations into continuous vector space, has quickly gained significant attention \cite{nickel2011three,lin2015learning,bordes2013translating,yang2014learning,trouillon2016complex}. 
These compact embeddings can preserve the inherent characteristics of entities and relations while enabling the use of these knowledge facts for a large variety of downstream tasks such as link prediction, question answering, and recommendation.

Despite the increasing success and popularity of Knowledge graph embeddings, their robustness has not been fully analyzed. In fact, many knowledge graphs are built upon unreliable or even public data sources. For instance, the well known \emph{Freebase} harvests its data from various sources including individual, user-submitted wiki contributions\footnote{https://www.nytimes.com/2007/03/09/technology/09data.html}. The openness of such data unfortunately would make KGE vulnerable to malicious attacks. When being attacked, substantial unreliable or even biased knowledge graph embeddings would be generated, leading to serious impairment and financial loss of many downstream applications. 
For instance, a variety of recommendation algorithms (e.g.,~\cite{zhang2016collaborative,wang2018dkn}) utilize KGEs of
products as external references. If KGEs are manipulated, the recommendation
results will be biased. This phenomenon can largely hurt user experiences.
Therefore, there is a strong need for the analysis of the vulnerability of knowledge graph embeddings. 

In this paper, for the first time, we systemically investigate the vulnerability of KGE, through designing efficient adversarial attack strategies. Due to the unique characteristics of knowledge graph and its embedding models, existing adversarial attack methods on graph data~\cite{zugner2018adversarial,sun2018data,bojcheski2018adversarial}
cannot be directly applied to attack KGE methods.
First,  they are all designed for homogeneous graphs, in which there is only a single type of nodes or links. However, in a knowledge graph, both the entities (nodes) and the relations (links) between entities are of different types. 
Second, existing attack methods for homogeneous graphs usually have strict requirements on the formulation of the targeted methods. For instance, the attack strategies proposed in \cite{sun2018data,bojcheski2018adversarial} can only work for the embedding methods that can be transformed into matrix factorization. However, the KGE methods are diverse and may not be able to be transformed into matrix factorization problems.
 
In this paper, we introduce the first study on the vulnerability of KGE and propose a family of effective \emph{data poisoning attack} strategies against KGE methods.
Our proposed attack strategies can guide the adversary to manipulate the training set of KGE by adding and/or deleting some specific facts to promote or degrade the plausibility of specific targeted facts, which can potentially influence a large variety of applications that utilize the knowledge graph.
The proposed strategies include both direct scheme which directly manipulates the embeddings of entities involved in the targeted facts and indirect scheme which utilizes other entities as proxies to achieve the attack goal.
Empirically, we perform poisoning attack experiments against three most representative KGE methods on two common KGE datasets (FB15K, WN18), and verify the effectiveness of the proposed adversarial attack.
Results show that the proposed strategies can dramatically worsen the link prediction results of targeted facts with only a small amount of changes to the graph needed. 

\section{Related Work}

\textbf{Knowledge Graph Embeddings}. KGE as an emerging research topic has attracted tremendous interest. A large number of KGE models have been proposed to represent entities and relations in a
knowledge graph with vectors or matrices. 
RESCAL~\cite{nickel2011three}, which is based on bi-linear matrix factorization, is one of the earliest KGE models. Then \cite{bordes2013translating} introduces the first translation-based KGE method TransE.
Given a fact $(h, r, t$), composed of a relation ($r$) and two entities ($h$ and $t$) in the knowledge graph, TransE learns vector representations of $h$, $t$, and $r$ (i.e., $\textbf{h}$, $\textbf{t}$ and $\textbf{r}$) by compelling $\textbf{h} + \textbf{r} \approx \textbf{t}$.
Later, a large collection of variants, such as TransH~\cite{wang2014knowledge},
TransR~\cite{lin2015learning}, TransD~\cite{ji2015knowledge} and TransA~\cite{xiao2015transa}, extend TransE by projecting the embedding vector into various spaces.
On the other hand, DistMult~\cite{yang2014embedding} simplifies RESCAL by only using a diagonal matrix, and ComPlex~\cite{trouillon2016complex} extends DistMult into the complex number field. 
\cite{wang2017knowledge} provides a comprehensive survey on these models.
The attack strategy proposed in this paper can be used to attack most of the existing KGE models. 

\textbf{Data Poisoning Attack v.s. Evasion Attack}.
\emph{Data poisoning attacks}, such as those in \cite{biggio2012poisoning,steinhardt2017certified} are a family of adversarial attacks on machine learning methods.
In these works, the attacker can access the training data of the learning algorithm, and has the power to manipulate a fraction of the training data in order to make the trained model meet certain desired objectives. 
\emph{Evasion attacks} such a those in \cite{goodfellow2014explaining,kurakin2016adversarial} are another prevalent type of attack that may be encountered in adversarial settings. In the evasion setting, malicious samples are generated at test time to evade detection. 
{In this paper, the proposed adversarial attack strategies against KGE methods can be categorized into the  \emph{data poisoning attack }}setting.

\textbf{Adversarial Attacks on Graphs} 
There are limited existing works on adversarial attacks for graph learning tasks: node classification~\cite{zugner2018adversarial,dai2018adversarial}, graph classification~\cite{dai2018adversarial}, link prediction~\cite{chen2018fast} and node embedding~\cite{sun2018data,bojcheski2018adversarial}. The first work, introduced by \cite{zugner2018adversarial} linearizes the graph convolutional network (GCN)~\cite{kipf2016semi} to derive the closed-form expression for the change in class probabilities for a given edge/feature perturbation and greedily pick the top perturbations that change the class probabilities. 
\cite{dai2018adversarial} proposes a reinforcement learning based  approach where the attack agent interacts with the targeted graph/node classifier to learn the policy of selecting the edge perturbations that fool the classifier. \cite{chen2018fast}
adopts the fast gradient sign scheme to perform \emph{evasion attack} against the link prediction task with GCN. \cite{sun2018data} and \cite{bojcheski2018adversarial} propose \textit{data poisoning attack} against factorization-based embedding methods on homogeneous graphs. They both formulate the poisoning attack as bi-level optimization problems. The former exploits the eigenvalue perturbation theory~\cite{stewart1990matrix}, while the latter directly adopts iterative gradient method~\cite{carlini2017towards} to solve the problem. To the best of our knowledge, there is no existing investigation on adversarial attack for heterogeneous graphs, in which the links and/or nodes are of different types, like knowledge graphs. This paper sheds first light on this important problem that has not been studied yet.


\section{Data Poisoning Attack against Knowledge Graph Embedding (KGE) Methods}

Let us consider a knowledge graph $\mathcal{KG}$, with a training set denoted as $\{(e_{n}^{h}, r_n, e_{n}^{t})\}_{n=1}^N$
	and a targeted fact triple $(e_{x}^{h,target}, r_{x}^{target}, e_{x}^{t,target})$ that does not exist in the training set. The goal of the attacker is to manipulate the learned embeddings, which would \textit{degrade} (or \textit{promote}) the plausibility of $(e_{x}^{h,target}, r_{x}^{target}, e_{x}^{t,target})$ measured by a specific \emph{fact plausibility scoring} function $f$. 
Without loss of generality, we focus on \ul{\textit{degrading}} the targeted fact.
We also assume that the attacker has a limited attacking budget. \emph{In this paper, the attacking budget is the
number of perturbations per target}.
Formally, the attack task is defined as follows:
\begin{definition}[Problem Definition]
Consider a targeted fact triple $(e_{x}^{h,target}, r_{x}^{target}, e_{x}^{t,target})$ that does not exist in the training set,
	we use $\textbf{e}_{x}^{h,target}$ to denote the embedding of the head entity $e_{x}^{h,target}$, $\textbf{e}_{x}^{t,target}$ to denote the embedding of the tail entity $e_{x}^{t,target}$ and $\textbf{r}_{x}^{target}$ to denote the embedding of the relation $r_{x}^{target}$ from the original training set.
	Our task is to \textcolor{black}{minimize the plausibility of $(e_{x}^{h,target}, r_{x}^{target}, e_{x}^{t,target})$, i.e., $f(\textbf{e}_{x}^{h,target}, \textbf{r}_{x}^{target}, \textbf{e}_{x}^{t,target})$, }
	by making perturbations (i.e., adding/deleting facts) on the training set. We assume the attacker has a given, fixed budget and is only capable of making $M$ perturbations. 
\end{definition}
Due to the discrete and combinatorial nature of the knowledge graph, solving this problem is highly challenging. 
Intuitively, in order to manipulate the plausibility of a specific targeted fact, we need to shift either the embedding vectors related to its entities or the embedding vectors/matrices related to its relations.
However, in a knowledge graph, the number of facts that a relation type involves is much larger than the number of facts that an entity type involves.
For instance, in the well-known knowledge graph \textit{Freebase}, the number of entities is over 30 million, while the number of relation types is only 1345. 
This leads to the fact that the innate \textit{characteristics} of each relation type is far more stable than that of entities and is difficult to be manipulated via a small number of modifications. Hence, in this paper, we focus on manipulating the plausibility of targeted facts from the perspective of entities. To achieve the attack goal, in the rest of this section, we propose a collection of effective yet efficient attack strategies. 

\subsection{Direct Attack}\label{sec:d}


Given the uncontaminated knowledge graph, the goal of direct attack is to determine a collection of perturbations (i.e., fact adding/deleting actions) to shift the embeddings of the entities involved in the targeted fact to minimize the plausibility of the targeted fact. First, we determine the optimal shifting direction that the entity's embedding should move towards.
Then we rank the possible perturbation actions by analyzing the training process of KGE models and designing scoring functions, which estimate the benefit of a perturbation, i.e., how much shifting can be achieved by this perturbation along the desired direction. We name the score as \emph{perturbation benefit score} and calculate such score for every possible perturbation.
Finally, we conduct the Top-$M$ perturbations with highest perturbation benefit scores, where $M$ is the attack budget.

Suppose we want to {\emph{degrade}} the plausibility of the fact $(e_{x}^{h,target}, r_{x}^{target}, e_{x}^{t,target})$.  For simplicity, let's focus on shifting the embedding of one of the entities in $(e_{x}^{h,target}, r_{x}^{target}, e_{x}^{t,target})$, say head entity $e_{x}^{h,target}$, from $\bm{e}_{x}^{h,target}$ to $\bm{e}_{x}^{h,target} + \bm{\epsilon}^*_x$, without loss of generality. Here, $\bm{\epsilon}^*_x$ denotes the \emph{embedding shifting vector}.  
The fastest direction of decreasing $f(\bm{e}_{x}^{h,target}, \bm{r}_{x}^{target}, \bm{e}_{x}^{t,target})$ is opposite to its partial derivative with respect to $\bm{e}_{x}^{h,target}$.
Let $\epsilon_h$ be the perturbation step size, the optimal embedding shifting vector is: 
\begin{equation}
\bm{\epsilon}_x^* = -\epsilon_h \cdot \frac{ \partial f(\bm{e}_{x}^{h,target}, \bm{r}_{x}^{target}, \bm{e}_{x}^{t,target})}{\partial \bm{e}_{x}^{h,target} }.
\label{eq:perturb}
\end{equation}

As mentioned in the problem definition, in order to shift $\bm{e}_{x}^{h,target}$ by $\bm{\epsilon}_x^*$, the adversary is allowed to add perturbation facts to the knowledge graph or delete facts from the knowledge graph. Given the optimal embedding shifting vector $\bm{\epsilon}_x^*$, we then find a ranking of the all the perturbation (add or delete) candidates. We discuss the two schemes in detail as follows.

\textbf{Direct Deleting Attack}:
Consider the uncontaminated training set, under the direct adversarial attack scheme, in order to shift the embedding of $e_x^{h, target}$ to $\bm{e}_x^{h, target} + \bm{\epsilon}_x^*$, we need to select and delete one or more facts that directly involve entity $e_x^{h, target}$. 
Intuitively, the fact to delete should have a great influence on the embedding of $e_x^{h, target}$, while at the same time not hinder the process of shifting the embedding of $e_x^{h, target}$  to $\bm{e}_x^{h, target} + \bm{\epsilon}_x^*$.
To design a scoring criterion that captures these intuitions, let us look into the training process of KGE model. Consider the specific deletion candidate $(e_x^{h, target}, r_i, e_i^t)$ that involves $e_x^{h, target}$. During training, the sum of the fact plausibility scores of the observed training samples is maximized. On one hand, the more plausible the fact $(e_x^{h, target}, r_i, e_i^t)$ is, the more it contributes to the final embedding of $e_x^{h, target}$. Hence, the perturbation benefit score of deleting $(e_x^{h, target}, r_i, e_i^t)$ should be proportional to $f(\bm{e}_x^{h, target}, \bm{r}_i, \bm{e}_i^t)$. On the other hand, if the plausibility of fact $(e_x^{h, target}, r_i, e_i^t)$ is large after $\bm{e}_x^{h, target}$ is shifted to $\bm{e}_x^{h, target} + \bm{\epsilon}_x^*$ (i.e., $f(\bm{e}_x^{h, target} + \bm{\epsilon}_x^*, \bm{r}_i, \bm{e}_i^t)$ is large), it means that the fact $(e_x^{h, target}, r_i, e_i^t)$ has a great positive impact on the embedding shifting and should not be deleted.
Hence, the perturbation benefit score of deleting $(e_x^{h, target}, r_i, e_i^t)$ should be inversely proportional to $f(\bm{e}_x^{h, target} + \bm{\epsilon}_x^*, \bm{r}_{i}, \bm{e}_i^t)$.
Formally, let the set of all the delete candidates be: $\mathcal{D_D} = \{(e_i^{h}, r_i, e_i^{t}) \;|\; e_i^{h} = e_x^{h, target} \;\text{and}\; (e_i^{h}, r_i, e_i^{t})\in \mathcal{KG} \}$, which intuitively denote the set of facts that involve $e_x^{h, target}$ as the head entity in the training set.
The perturbation benefit score of deleting a specific perturbation fact $(e_x^{h, target}, r_i, e_i^t)$ can be estimated as:
\begin{equation}
\begin{split}
\eta^-(e_x^{h, target}, r_i, e_i^t) = & f(\bm{e}_x^{h, target}, \bm{r}_{i}, \bm{e}_{i}^t) \\
& - \lambda_1  f(\bm{e}_x^{h, target} + \bm{\epsilon}_x^*, \bm{r}_{i}, \bm{e}_{i}^t),
\end{split}
\label{eq:dd}
\end{equation}
where $\bm{e}_x^{h, target}$, $\bm{r}_{i}$, and $\bm{e}_i^t$ denote the embeddings of $e_x^{h, target}$, $r_i$ and $e_i^t$, respectively,  on the uncontaminated training set. 


\textbf{Direct Adding Attack}:
Now we discuss how to conduct 
direct adding perturbation.
To shift the embedding of $e_{x^{h, target}}$ by $\bm{\epsilon}^*$, we just need to add new facts that involve $e_x^{h, target}$ to make  $f(\bm{e}_x^{h, target}, \bm{r}_{j}, \bm{e}_{j}^t)$ less plausible. The set of all the possible  adding candidates can be denoted as $\mathcal{D_A} = \{e_x^{h, target}\} \times \{(r_j, e_j^{t}) \;|\; \forall r_j \in \mathcal{KG} \text{ and } e_j^{t}\in \mathcal{KG} \}$, where $\{(r_j, e_j^{t}) \;|\; \forall r_j \in \mathcal{KG} \text{ and } e_j^{t}\in \mathcal{KG} \}$ denotes \textcolor{black}{all the possible ``relation-tail entity'' combinations in the knowledge graph and $\times$ stands for Cartesian product}. In practice, for better efficiency, we can downsample a subset from all the possible ``relation-tail entity''  combinations.
Formally, the perturbation benefit score of a specific candidate to add (i.e., $(e_x^{h, target}, r_j, e_j^t)$) can be estimated as:
\begin{equation}
\begin{split}
\eta^+(e_x^{h, target}, r_j, e_j^t) = 
& \lambda_2 f(\bm{e}_x^{h, target} + \bm{\epsilon}_x^*, \bm{r}_{j}, \bm{e}_{j}^t)  \\ 
& - \lambda_3 f(\bm{e}_x^{h, target}, \bm{r}_{j}, \bm{e}_{j}^t),
\end{split}
\label{eq:da}
\end{equation}
where $\bm{e}^{h, target}_x$, $\bm{r}_{j}$, and $\bm{e}_{i}^t$ denote the embeddings on the uncontaminated training set. 


\subsection{Indirect Attack}\label{sec:ia}

Although the direct attack strategy is intuitive and effective, {it is possible to be detected by data sanity check.} 
In this section, we move on to introduce a more complicated yet more stealthy adversarial attack scheme, i.e., indirect attack.
Suppose a KGE user want to query the plausibility of a potential fact $(h, r, t)$. Due to the huge scales of real-world knowledge graphs, even in the most optimistic situation, we may merely carry data sanity test on the facts related to $h$ and $t$. However, for indirect attack, instead of adding or deleting the facts that involve the entities in the targeted fact, we propose to perturb the facts that involve other entities in the knowledge graph and let the perturbation effect propagate to the targeted fact. Thus,
detecting these perturbations requires data sanity tests on facts that involves every entity that are hops away from $h$ and $t$. When the number of hops increases linearly, the data sanity cost will have a exponential growth. Even though there is an Oracle that can find these anomalous facts effectively, defenders cannot determine the targeted fact(s) of these perturbations.
For a better description, we provide the following toy example, which is used throughout this section. 
\begin{example}
	Suppose we want to degrade the plausibility of the targeted fact $(e_{x}^{h,target}, r_{x}^{target}, e_{x}^{t,target})$ via shifting the embedding of the targeted entity $e_{x}^{h,target}$ by $\bm{\epsilon}_x^*$, without loss of generality. Under indirect attack scheme, we perturb the facts that involve the K-hop neighbors of $e_{x}^{h,target}$. These  K-hop neighbors are called \emph{proxy entities}. Then the entities between the K-hop neighbors (proxy entities) and $e_{x}^{h,target}$ are \emph{intermediate entities} to propagate the influence of the perturbations to $e_{x}^{h,target}$.
	The propagation path can be illustrated as follows:
	\begin{equation*}
	\begin{split}
	&\text{\textbf{Propagation Path}}:\\
	& e_x^{h,target} \xleftrightarrow{r_{x,1}} e_{x,1} \xleftrightarrow{r_{x,2}} e_{x,2} \cdots \xleftrightarrow{r_{x,K}} e_{x,K}
	\end{split}
	\end{equation*}
	where we use $\xleftrightarrow{r_{x,\cdot}}$ to denote the directional relation and use notation $e_{x,\cdot}$ to denote the entities on the path. A specific $e_{x,\cdot}$ can work as both the head entity and the tail entity. The notations in the path above are adopted in the rest of this section.
\end{example}
When the perturbations on the proxy entity cause an embedding shift on itself, the embeddings of its neighboring entities will also be influenced. 
The influence will propagate back to the embedding of the {targeted entity} ultimately.

However, finding the effective perturbations on the {proxy entities}, which are $K$-hop away from the targeted entity, is indeed a challenging task. The task involves two key problems: \textbf{(1)} \emph{Given a specific propagation path, how can we determine the desired embedding shifting vectors on its intermediate entities and its proxy entity, in order to accomplish the embedding shifting goal on the targeted entity?} \textbf{(2)} \emph{How do we select the propagation paths to propagate the influence of perturbation to the targeted entity?} 

In the rest of this section, we discuss strategies to solve these key problems and propose a criterion to evaluate the benefit of an indirect perturbation (i.e., the \emph{perturbation benefit score}).

\emph{For the first problem}, given a specific path, in order to conduct a perturbation that makes the embedding of $e_{x}^{h,target}$ shift towards the desired direction (i.e., the direction of $\bm{\epsilon}_x^*$), we decide the shifting goal for each entity on the path in a \emph{recurrent} way.
Suppose we want to shift $\bm{e}_{x}^{h,target}$ by $\bm{\epsilon}_x^{*}$ via the intermediate entities along the path specified in Example~1. The entity that directly influences $\bm{e}_{x}^{h,target}$ is its neighbor $e_{x,1}$ and what we need to do is to determine the ideal embedding shifting vector $\bm{\epsilon}_{x,1}^{*}$ on $e_{x,1}$, so that the desired embedding shift on $e_x^{h,target}$ (i.e., $\bm{\epsilon}_x^*$) is approached to the greatest extent. 
Formally, $\bm{\epsilon}_{x,1}^{*}$ should satisfy:
\begin{equation}
\begin{split}
\bm{\epsilon}_{x,1}^{*} = \arg\max_{\bm{\epsilon}} \; &
f(\bm{e}_{x}^{h,target} + \bm{\epsilon}_x^*, \bm{r}_{x,1}, \bm{e}_{x,1}  + \bm{\epsilon}) \\
& - f(\bm{e}_{x}^{h,target}, \bm{r}_{x,1}, \bm{e}_{x,1}  + \bm{\epsilon})  
\\ 
s.t. \; & \textcolor{black}{||\bm{\epsilon}||_2} = \epsilon_h,
\end{split}
\label{eq:perturb_t}
\end{equation}
where $\epsilon_h$ is the perturbation step size, $\bm{e}_{x,1}$ denotes the embedding of ${e}_{x,1}$, and $\bm{r}_{x,1}$ denotes the embedding of ${r}_{x,1}$. As a result, the embedding of $e_{x}^{h,target}$ will have a larger tendency to move towards $\bm{e}_{x}^{h,target} + \bm{\epsilon}_x^*$ than towards $\bm{e}_{x}^{h,target}$, during the training process on the contaminated training data.
When $\bm{\epsilon}_{x,1}^{*}$ is determined, we can further get the embedding shifting vector for $e_{x,2}, \cdots, e_{x,K}$, which are denoted as $\bm{\epsilon}_{x,2}^*, \cdots, \bm{\epsilon}_{x,K}^*$, respectively. This process is similar as above.

With the embedding shifting vectors on the proxy entities of each path determined, we calculate the scores $\eta^-$ and $\eta^+$, defined in Eq.~\eqref{eq:dd} and \eqref{eq:da} for all the possible add/delete perturbations. These scores are later used to calculate the perturbation benefit score under indirect attack schemes.


\begin{algorithm}[t!]
	\caption{Indirect Attack}
	\label{alg}
	\begin{algorithmic}[1]
		\Require Targeted fact $(e_{x}^{h,target}, r_{x}^{target}, e_{x}^{t,target})$, Neighbor hop $K$, Targeted entity $e_{x}^{h,target}$.
		\State Exhaust all the $K$-hop paths originating from $e_{x}^{h,target}$
		\State Exhaust all the possible perturbation candidates on the proxy entities of these $K$-hop paths.
		\State Calculate  $\bm{\epsilon}_x^*$ for $e_{x}^{h,target}$ according to Eq.~\eqref{eq:perturb}.
		\For {each path $k$}
		\For {each intermediate entity / proxy entity $e_{x,i}$ in path $x$}
		\State Calculate $\bm{\epsilon}_{x,i}^*$ according to Eq.~\eqref{eq:perturb_t}.
		\EndFor
		\State Calculate the score $\psi$ for each perturbation on the current proxy entity according to Eq.~\eqref{eq:dd}, Eq.~\eqref{eq:da} and \eqref{eq:score}.
		\EndFor
		\State Select $M$ perturbations with highest scores (i.e., $\psi$) and conduct the attack.
	\end{algorithmic}
\end{algorithm}

\emph{For the second problem}, we look into the training objective function. Suppose we want to shift the embedding of $e_{x,k-1}$ via its neighbor $e_{x,k}$, when the embedding shift on $\bm{e}_{x,k}$ is $\bm{\epsilon}_{x,k}^*$. 
To estimate the influence of such embedding shift on $\bm{e}_{x,k-1}$, we isolate all the facts that involve $e_{x,k-1}$ in the training objective function, force a embedding shift $\bm{\epsilon}_{x,k}$ on $\bm{e}_{x,k}$ and ignore the negative sampling terms. Formally, the objective function becomes: $
\min_{\bm{e}_{e_{x,k-1}}} 
\sum_{(e_i^h,r_i,e_i^t)\in D_{e_{x,k-1}}^{\setminus e_{x,k}}} \mathcal{L}(\bm{e}_{i}^h,\bm{r}_{i},\bm{e}_{i}^t)  + \mathcal{L}(\bm{e}_{x,k-1},\bm{r}_{x,k},\bm{e}_{x,k}+ \bm{\epsilon}_{x,k})$,
where $D_{e_{x,k-1}}^{\setminus e_{x,k}}$ stands for the set of all the observed facts, which involve $e_{x,k-1}$ except the fact $(e_{x,k-1}, r_{x,k}, e_{x,k})$, in the training set. $\mathcal{L}$ denotes the loss function for a single fact. $\bm{e}_{x,k}+ \bm{\epsilon}_{x,k}$ in $\mathcal{L}(\bm{e}_{x,k-1},\bm{r}_{x,k},\bm{e}_{x,k}+ \bm{\epsilon}_{x,k})$ indicates that the embedding of $e_{x,k}$ is already shifted. \textcolor{black}{Clearly, if we fix the embeddings of all the relations and entities except $e_{x,k-1}$, the impact of shifting $\bm{e}_{x,k}$ to $\bm{e}_{x,k}+ \bm{\epsilon}_{x,k}$ is highly correlated with the number of facts that involves $e_{x,k-1}$, i.e., $|D_{e_{x,k-1}}|$. That is to say, the more neighbors an entity has, the \emph{less} it will be influenced by a specific perturbation on one of its neighbors. }

Based on  above discussions, we propose an empirical scoring function to evaluate the perturbation benefit score of every possible perturbation. We still consider the scenario specified in Example~1. Suppose we conduct an add/delete perturbation $(e_{x,K}, r_{x,K}, e_{x,K+1})$ on the proxy entity $e_{x,K}$. The perturbation benefit score of this indirect perturbation is defined as:
\begin{equation}
\begin{split}
& \psi(e_{x,K}, r_{x,K}, e_{x,K+1}) \\
= \;& \eta(e_{x,K}, r_{x,K}, e_{x,K+1})
- \lambda \log \big(\frac{1}{K}\sum_{k=1}^{K-1} |D_{e_{x,k-1}}| \\
& +  \max(\{ |D_{e_{x,k-1}}| \}_{k=1}^{K-1}) \big),
\end{split}
\label{eq:score}
\end{equation}
where $\max(\{ |D_{e_{x,k-1}}| \}_{k=1}^{K-1})$ stands for the maximum number of facts that involves each entity $k$ on the path. $\eta$ is the same as $\eta^+$ under add perturbation scheme and is the same as $\eta^-$ under delete perturbation scheme. $\lambda$ is a trade-off parameter. 
The first term estimates the direct perturbation benefit of the perturbation in terms of shifting the proxy entity as desired.
The second term evaluates the capability of the intermediate entities on the path in terms of propagating the influence to the targeted entity. As the influence may be diluted by the facts that involve each entity $e_{x,k}$ on the path. A smaller averaged number of facts that involves each entity $e_{x,k}$ on the path indicates a larger capability of the path in terms of propagating the influence. Moreover, we also consider the maximum number of facts that involves each entity $e_{x,k}$ on the path. This is to avoid the case when some intermediate entities, whose embedding is difficult to shift, ``block'' the propagation path. \emph{In practices, we can first utilize the second term to determine the best $P$ paths in terms of propagating the influence from proxy entities to  the targeted entity and then choose what facts to add or delete upon these proxy entities in the best $P$ paths.~\footnote{This strategy is used in the experiments of this paper.}}
The overall workflow of indirect attack is illustrated in Algorithm~\ref{alg}.

%

\section{Experiments}

In this section, we evaluate the proposed attack strategies under different settings on two benchmark datasets. 

\subsection{Datasets and  Settings}

\textbf{Datasets}:
In this paper, we use two common KGE benchmark datasets for our experiment:  FB15k and WN18. FB15k is a subset of Freebase, which is a large collaborative knowledge base consisting of a large number of real-world facts. WN18 is a subset of Wordnet~\footnote{https://wordnet.princeton.edu/}, which is a large lexical knowledge graph.
Both FB15k and WN18 are first introduced by~\cite{bordes2013translating}.
The training set and the test set of these two datasets are already fixed. \emph{We randomly sample 100 samples in the test set as the targeted facts for the proposed attack strategies.}


\textbf{Baseline \& Targeted Models}:
Since there are no existing methods that can work under the setting of this paper, we compare the proposed attack schemes with several naive baseline strategies. Specifically, we design \emph{random-dd} (random direct deleting), \emph{random-da} (random direct adding), \emph{random-id} (random indirect deleting), \emph{random-ia} (random indirect adding) as comparison baselines for our proposed \emph{direct deleting attack}, \emph{direct adding attack}, \emph{indirect deleting attack}, \emph{indirect adding attack}, respectively. 
{The difference between the baseline and its corresponding proposed methods is that the perturbation facts to add/delete are randomly selected.}
For the targeted KGE models, we choose three most representative \textit{TransE}~\cite{bordes2013translating}, \textit{TransR}~\cite{lin2015learning} and \textit{RESCAL}~\cite{nickel2011three} as attack targets.

\textbf{Metrics}: In order to evaluate the effectiveness of the proposed attack strategies. 
We compare the plausibility change of the targeted fact before and after the adversarial attack.
Specifically, we follow the evaluation protocol of KGE models described in the previous works like~\cite{bordes2013translating}. Given a targeted fact $(e_h, r, e_t)$, we remove the head or tail entity and then replace it with all the possible entities. 
We first compute plausibility scores of those corrupted facts and then rank them by descending order; the rank of the correct entity is stored.
After that, we use \emph{MRR} (Mean Reciprocal Rank of all the ground truth triples) and \emph{H@10} (the proportion of correct entities ranked in top 10, for all the ground truth entities.) as our evaluation metrics. \emph{The smaller MRR and H@10 are on the contaminated dataset, the better the attack performance is.}

\textbf{Experiment Settings}:
For the targeted KGE models, we use the standard implementation provided by THUNLP-OpenKE~\footnote{https://github.com/thunlp/OpenKE}~\cite{han2018openke}.
The embedding dimension $d$ is fixed to 50. Other parameters of baseline methods are set according to their authors' suggestions. For the proposed attack strategies, the parameter $K$ for indirect attack is fixed to 1. During the experiment all the perturbations are injected into the dataset at the same time.
The attack models in this paper are all implemented via Numpy and Python 3.7. The attack models are run on a laptop with 4 GB RAM, 2.7 GHz Intel Core i5 CPU. 

\subsection{Results and Analysis}

In this section, we report and analyze the attack results of the proposed attack strategies under different settings. To avoid confusion, the performance of direct adding attack, direct deleting attack, indirect adding attack, and indirect deleting attack are reported separately in Table~\ref{table:da},~\ref{table:dd},~\ref{table:ia} and~\ref{table:id}.

\begin{table}[t]\scriptsize
\centering
	\begin{tabular}{llllllll}
		\toprule
		&  &\multicolumn{2}{c}{Clean} & \multicolumn{2}{c}{random-da} & \multicolumn{2}{c}{Direct Add}  \\
		&  & MRR & H@10 & MRR & H@10 & MRR & H@10 \\
		\hline
		& TransE & 0.26  & 0.49 & 0.24 & 0.46 & {0.24} & {0.42} \\
		FB15K & TransR & 0.24 & 0.52 & 0.23 & 0.42 & {0.21} & {0.41}  \\
		& RESCAL & 0.19 & 0.42 & 0.20 & 0.40 & {0.17} & {0.39} \\
		\hline
		& TransE & 0.39 & 0.70 & 0.30 & 0.68 & {0.21} & {0.53} \\
		WN18 & TransR & 0.44 & 0.73 & 0.41 & 0.71 & {0.22} & {0.51} \\
		& RESCAL & 0.41 & 0.72 &  0.44 & 0.69 & {0.30} & {0.57} \\
		\bottomrule
	\end{tabular}
	\caption{Overall Results of Direct Adding Attack}
	\label{table:da}
\end{table}

\begin{table}[t]\scriptsize
\centering
	\begin{tabular}{llllllll}
		\toprule
		&  &\multicolumn{2}{c}{Clean} & \multicolumn{2}{c}{random-dd} & \multicolumn{2}{c}{Direct Delete}  \\
		&  & MRR & H@10 & MRR & H@10 & MRR & H@10 \\
		\hline
		& TransE & 0.26  & 0.49 & 0.26 & 0.54 & {0.19} & {0.37} \\
		FB15K & TransR & 0.24 & 0.52 & 0.25 & 0.49 & {0.18} & {0.41} \\
		& RESCAL & 0.19 & 0.42 & 0.19 & 0.38 & {0.13} & {0.30} \\
		\hline
		& TransE & 0.39 & 0.70 & 0.36 & 0.71 & {0.11} & {0.26} \\
		WN18 & TransR & 0.44 & 0.73 & 0.43 & 0.68 & {0.11} & {0.24} \\
		& RESCAL & 0.41 & 0.72 & 0.40 & 0.67 & {0.02} & {0.05} \\
		\bottomrule
	\end{tabular}
	\caption{Overall Results of Direct Deleting Attack}
	\label{table:dd}
\end{table}

\textbf{Overall Attack Performance}: Let us first discuss the performances of the direct attack schemes on two datasets. For the direct deleting attack scheme, we set the attack budget for each targeted fact to 4 and 1 on FB15K and WN18 dataset, respectively. 
For the direct attacking attack scheme, the attack budgets for each targeted fact are 8 and 6 for FB15K and WN18 dataset, respectively.
\textcolor{black}{These budgets are low enough to make the whole attack process unnoticeable.} From the results, we can clearly see that the plausibilities of these targeted facts significantly degrade as desired.  We can conclude that these KGE models are quite vulnerable to even a small number of perturbations generated by well-designed attack strategies. For comparison, we have also tested the baseline methods \emph{random-da} and \emph{random-dd}, which cannot achieve satisfactory attack performances.
This demonstrates the effectiveness of the proposed strategies. Moreover, we observe that the effectiveness of the proposed strategies is more significant on WN18 dataset than on FB15K dataset. This is because the average number of facts that each entity involves in WN18 dataset is significantly smaller than that in FB15K dataset. Hence, the graph structure of FB15K is more stable and robust.
Then, let us move on to the discussion of indirect attack schemes.
For the indirect adding attack, we set the attack budget for each targeted fact to 60 and 20 for FB15K and WN18 dataset, respectively.
For the indirect deleting attack, the attack budgets for each targeted fact are set to 20 and 5 for FB15K and WN18 dataset, respectively. 
The reason why indirect attacks need more attack budgets to get comparable results is that only a small portion of the influence caused by the perturbations on proxy entities is propagated to the targeted entity.
 In contrast, nearly all of the influence of the perturbation is exerted on the targeted entity under direct attack schemes.
Like direct attack schemes, these indirect attack schemes also demonstrate their effectiveness. For instance, under the indirect deleting attack scheme, the H@10 and MRR metrics of the targeted facts decrease by approximate 0.03 on FB15K dataset. Thus, the indirect deleting attack schemes can also be used in practices to make the attack process more stealthy.  

\begin{table}[t]\scriptsize
\centering
	\begin{tabular}{llllllll}
		\toprule
		&  &\multicolumn{2}{c}{Clean} & \multicolumn{2}{c}{random-ia} & \multicolumn{2}{c}{Indirect Add}  \\
		&  & MRR & H@10 & MRR & H@10 & MRR & H@10 \\
		\hline
		& TransE & 0.26  & 0.49 & 0.25 & 0.50 & {0.23} & {0.47} \\
		FB15K & TransR & 0.24 & 0.52 & 0.25 & 0.51 & {0.22} & {0.49} \\
		& RESCAL & 0.19 & 0.42 & 0.19 & 0.40 & {0.17} & {0.36} \\
		\hline
		& TransE & 0.39 & 0.70 & 0.42 & 0.71 & {0.32} & {0.67} \\
		WN18 & TransR & 0.44 & 0.73 & 0.40 & 0.73 & {0.34} & {0.69} \\
		& RESCAL & 0.41 & 0.72 & 0.41 & 0.69 & {0.39} & {0.63} \\
		\bottomrule
	\end{tabular}
	\caption{Overall Results of Indirect Adding Attack}
	\label{table:ia}
\end{table}

\begin{table}[t]\scriptsize
\centering
	\begin{tabular}{llllllll}
		\toprule
		&  &\multicolumn{2}{c}{Clean} & \multicolumn{2}{c}{random-id} & \multicolumn{2}{c}{Indirect Delete}  \\
		&  & MRR & H@10 & MRR & H@10 & MRR & H@10 \\
		\hline
		& TransE & 0.26  & 0.49 & 0.27 & 0.50 & {0.22} & {0.44} \\
		FB15K & TransR & 0.24 & 0.52 & 0.25 & 0.53 & {0.21} & {0.48} \\
		& RESCAL & 0.19 & 0.42 & 0.20 & 0.36 & {0.16} & {0.34} \\
		\hline
		& TransE & 0.39 & 0.70 & 0.44 & 0.74 & {0.35} & {0.68} \\
		WN18 & TransR & 0.44 & 0.73 & 0.45 & 0.74 & {0.41} & {0.71} \\
		& RESCAL & 0.41 & 0.72 & 0.42 & 0.70 & {0.38} & {0.64} \\
		\bottomrule
	\end{tabular}
	\caption{Overall Results of Indirect Deleting Attack}
	\label{table:id}
\end{table}

\begin{figure}[t]
 	\centering
 	\begin{subfigure}[b]{0.23\textwidth}
 		\includegraphics[width=\textwidth]{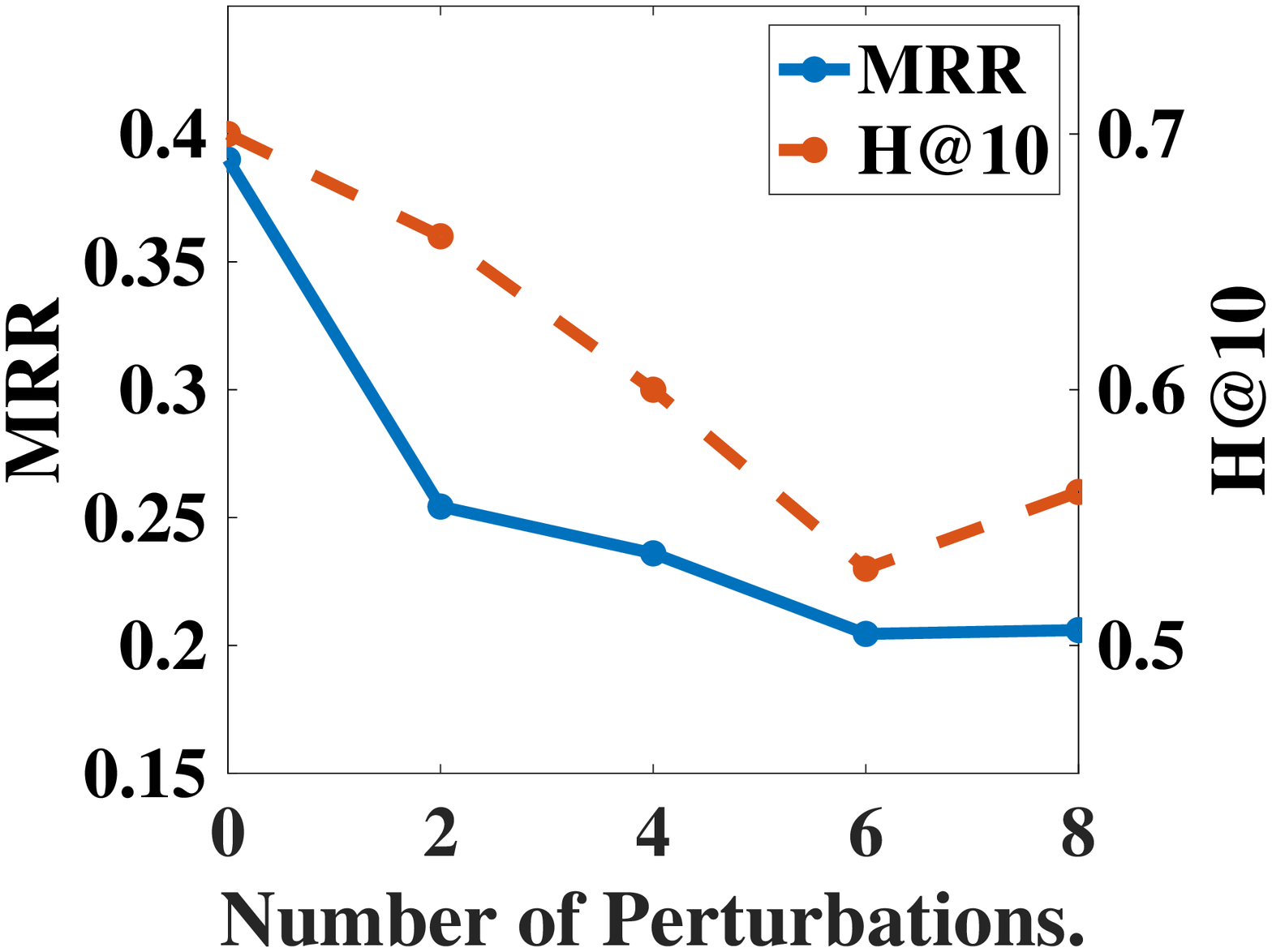}
 		\caption{Direct Adding Attack on WN18 Dataset against TransE}
 	\end{subfigure}
 	~
 	\begin{subfigure}[b]{0.22\textwidth}
 		\includegraphics[width=\textwidth]{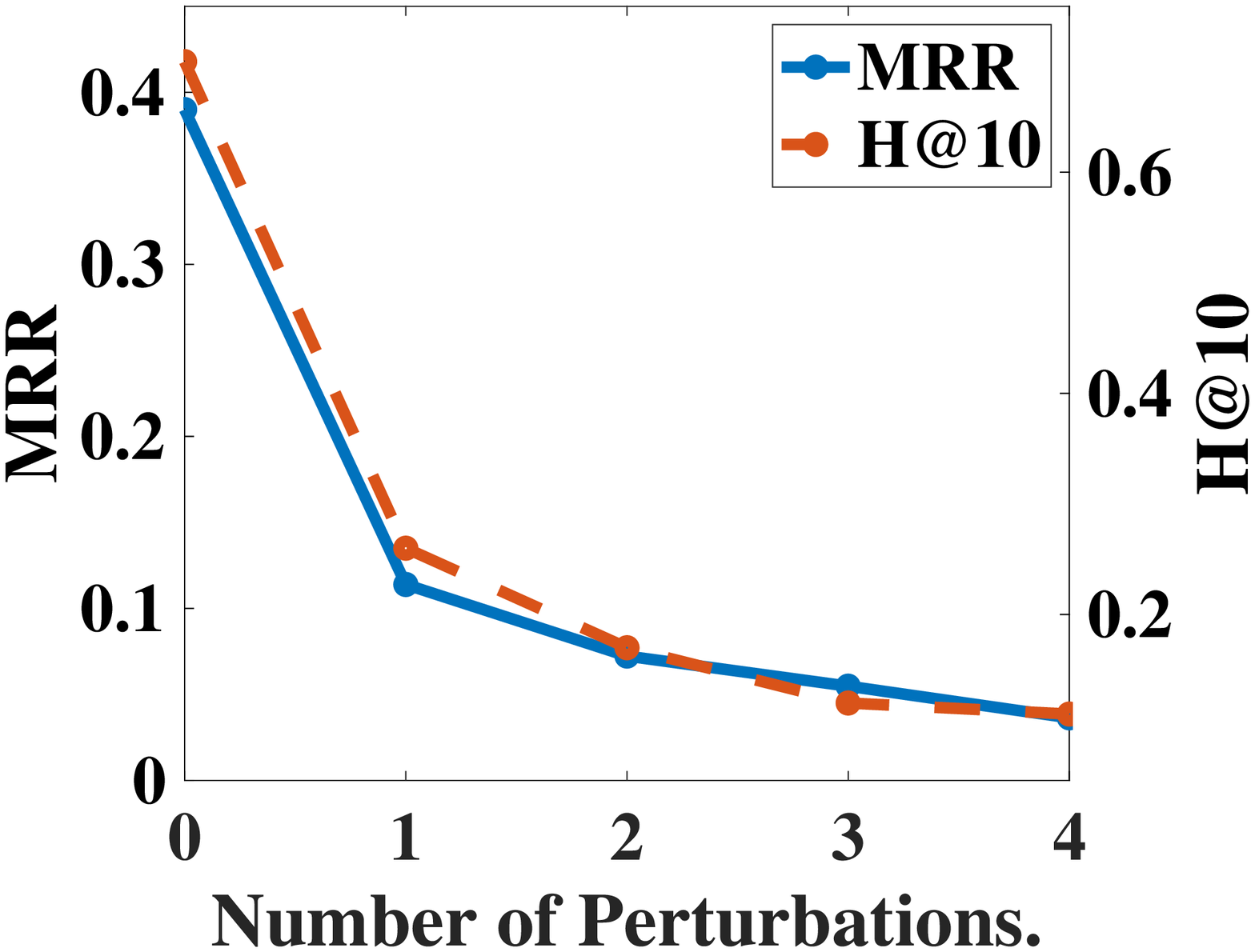}
 		\caption{Direct Deleting Attack on WN18 Dataset against TransE}
 	\end{subfigure}
 	\caption{Analysis of the Number of Perturbations}\label{fig:trans}
 \end{figure}

 \textbf{Analysis of the Number of Perturbations}: When conducting the data poisoning attack, one of the most important factors is the number of perturbations (i.e. attack budget). Due to space limit, we merely plot performances of direct  attack schemes against TransE w.r.t. the number of perturbations (i.e., attack budgets) on WN18 dataset in Figure~\ref{fig:trans}. From Figure.~\ref{fig:trans}, we can clearly see that the proposed attack strategies consistently degrade the plausibility of the targeted facts under both setting.  When the number of perturbations keeps increase, the growth of attack performance becomes slower.
 This is because when the number of perturbations is small, the selected perturbations are usually of high value in terms of manipulating the plausibility of the targeted facts. When the number of perturbations keeps increase, the high-value perturbations are used up. Hence, the performances become stable.

\textbf{Efficiency Analysis}: Finally, let us discuss the efficiency of the proposed attack strategies. Here we report the time consumption for the proposed attack strategies to generate the perturbations for a single targeted fact on average. The time consumption of  Direct Adding, Direct Deleting, Indirect Adding and Indirect Deleting scheme are 3.36s, 0.13s, 14.04s, and 1.22s, respectively.~\footnote{Note: Candidate downsampling, which is described in Section 3.1 is used.}
As one can see, the proposed model takes less than 15 seconds on average to generate the perturbations for a single targeted fact. For the direct deleting attack scheme, the time cost is less than 1 second on average. 
These results show that the proposed attack strategies are quite efficient.

\section{Conclusions}

We present the first study on the vulnerability of existing KGE methods and propose a collection of data poisoning attack strategies for different attack scenarios. These attack strategies can be efficiently computed. 
Experiment results on two benchmark dataset demonstrate that the proposed strategies  can effectively manipulate the plausibility of arbitrary facts in the knowledge graph with limited perturbations. 

\section*{Acknowledgments}
We thank our anonymous reviewers for their insightful comments and suggestions on this paper. This work was supported in part by the US National Science Foundation under grants CNS-1742847, IIS-1747614 and CNS-1652503.


\bibliographystyle{named}
\bibliography{ijcai19}

\end{document}